\documentclass[10pt,twocolumn,letterpaper]{article}

\usepackage[pagenumbers]{cvpr} % To force page numbers, e.g. for an arXiv version
\usepackage{graphicx}
\usepackage{amsmath}
\usepackage{amssymb}
\usepackage{booktabs}
\usepackage{array,multirow}
\usepackage{makecell}

\usepackage{stmaryrd}
\usepackage{comment}
\usepackage{amsmath,amssymb} % define this before the line numbering.
\usepackage{color}
\usepackage{bbm}
\usepackage{romannum}
\usepackage{xcolor}
\usepackage{caption}
\usepackage{graphicx}
\usepackage{enumitem}

\newcommand{\cmk}{\checkmark}

\newcommand*\samethanks[1][\value{footnote}]{\footnotemark[#1]}

\usepackage[pagebackref,breaklinks,colorlinks]{hyperref}

\usepackage[capitalize]{cleveref}
\crefname{section}{Sec.}{Secs.}
\Crefname{section}{Section}{Sections}
\Crefname{table}{Table}{Tables}
\crefname{table}{Tab.}{Tabs.}

\def\cvprPaperID{2484} % *** Enter the CVPR Paper ID here
\def\confName{CVPR}
\def\confYear{2023}

\begin{document}

%\title{SDBPS: Symmetric Distillation and Bilateral-consensus Pseudo Supervision for Domain Adaptive Semantic Segmentation}
\title{DiGA: \textit{Di}stil to \textit{G}eneralize and then \textit{A}dapt for Domain Adaptive\\Semantic Segmentation}

\author{Fengyi Shen\textsuperscript{1,2,3}, Akhil Gurram\textsuperscript{2}, Ziyuan Liu\textsuperscript{2}, He Wang\textsuperscript{3}\thanks{corresponding author}, Alois Knoll\textsuperscript{1}\samethanks\\
\noindent\textsuperscript{1}Technical University of Munich, \textsuperscript{2}Huawei Munich Research Center, \textsuperscript{3}EPIC Lab, Peking University\\
{\tt\small \textsuperscript{1}fengyi.shen@tum.de,knoll@in.tum.de,\textsuperscript{2}\{first.last\}@huawei.com,\textsuperscript{3}hewang@pku.edu.cn}
}
\maketitle

%%%%%%%%% ABSTRACT
\begin{abstract}
%Domain adaptive semantic segmentation aims to benefit from labelled source domain data and find the correct class-wise assignment for each pixel of unlabelled target domain images. %Stage-wise training has become a popular strategy in domain adaptation as it ensures stable learning to obtain robust cross-domain knowledge. However, current adversarial training for warm-up stage results in limited performance gain, and finding proper categorical thresholds is a challenging task in the following self-training stage.   
Domain adaptive semantic segmentation methods commonly utilize stage-wise training, consisting of a warm-up and a self-training stage. However, this popular approach still faces several challenges in each stage: for warm-up, the widely adopted adversarial training often results in limited performance gain, due to blind feature alignment; for self-training, finding proper categorical thresholds is very tricky.
To alleviate these issues, we first propose to replace the adversarial training in the warm-up stage by a novel symmetric knowledge distillation module that only accesses the source domain data and makes the model domain generalizable.
Surprisingly, this domain generalizable warm-up model brings substantial performance improvement, which  
can be further amplified via our proposed cross-domain mixture data augmentation technique.
Then, for the self-training stage,  we propose a threshold-free dynamic pseudo-label selection mechanism to ease the aforementioned threshold problem and make the model better adapted to the target domain. 
Extensive experiments demonstrate that our framework achieves remarkable and consistent improvements compared to the prior arts on popular benchmarks.
Codes and models are available at {\href{https://github.com/fy-vision/DiGA}{https://github.com/fy-vision/DiGA}}
%\keywords{Domain Adaptation, Semantic Segmentation, Unsupervised Learning, Autonomous Driving}
\end{abstract}

\section{Introduction}
\label{sec:intro}
 Semantic segmentation~\cite{long2015fully,chen2017deeplab,zhao2017pyramid,lin2017refinenet} is an essential component in autonomous driving~\cite{feng2020deep}, image editing~\cite{park2019semantic,zhang2020cross}, medical imaging~\cite{ronneberger2015u}, etc. However, for images in a specific domain, training deep neural networks~\cite{rumelhart1986learning,lecun1998gradient,krizhevsky2012imagenet,lecun2015deep} for semantic segmentation often requires a vast amount of pixel-wisely annotated data, which is expensive and laborious. Therefore, \textbf{domain adaptive semantic segmentation}, \textit{i.e.} learning semantic segmentation from a labelled source domain (either virtual data or an existing dataset) and then performing  unsupervised domain adaptation (UDA)~\cite{ben2007analysis,daume2009frustratingly,ganin2015unsupervised,long2016unsupervised,saito2018maximum} to the target domain,  becomes an important research topic. Yet the remaining challenge is the severe model performance degradation caused by the visual domain gap between source and target domain data. In this work, we tackle this domain adaptive semantic segmentation problem, with the goal of aligning correct categorical information pixel-wisely from the labelled source domain onto the unlabelled target domain.

Currently, stage-wise training, composed of a warm-up and a self-training stage, has been widely adopted in domain adaptive semantic segmentation~\cite{zheng2021rectifying, zhang2019category, zhang2021prototypical, li2019bidirectional, mei2020instance, araslanov2021self} as it stabilizes the domain adaptive learning~\cite{zhang2019category} and thus reduces the performance drop across domains.
However, the gap is far from being closed and, in this work, we identify that there is still a large space to improve in both warm-up and self-training.\\
\indent Regarding {\bf warm-up}, recent  works in this field~\cite{zheng2021rectifying, zhang2019category, zhang2021prototypical, li2019bidirectional, mei2020instance} mostly adopt adversarial training~\cite{ganin2015unsupervised,tzeng2017adversarial,tsai2018learning} as their basic strategy, which usually contributes to limited adaptation improvements.
Without knowing the target domain labels, this adversarial learning proposes to align the overall feature distributions across domains. 
Note that this alignment is class-unaware and fails to guarantee the features from the same semantic category are well aligned between the source and target domain, thus being sub-optimal as a warm-up strategy. 
%due to the blind and class-unaware alignment of target features towards source distribution, thus hampering further training. 

In contrast, %we propose {\bf not} to align features between the two domains for warm-up, and instead focus on first enhancing the model's domain generalizability even without the attendance of target data. 
we take an alternative perspective to improve the warm-up: simply enhancing the model's domain generalizability without considering target data.
To be specific, we introduce a pixel-wise symmetric knowledge distillation technique. The benefits are threefold: {\bf\textit{\romannum{1}}}.~knowledge distillation is performed on the source domain where ground-truths are available, the learning thus becomes class-aware, which avoids the blind alignment as observed in adversarial training; {\bf\textit{\romannum{2}}}.~the soft labels created in the process of distillation can effectively avoid the model overfitting to domain-specific bias~\cite{wang2021embracing} and help to learn more generalized features across domains; 
{\bf\textit{\romannum{3}}}.~our symmetric proposal ensures the bidirectional consistency between the original source view and its augmented view, leading to more generalizable model performance in the warm-up stage. Our method achieves a significant improvement from 45.2 to 48.9 mIoU compared to adversarial training, meanwhile, it outperforms existing arts on domain generalized segmentation.

We further observe that making the data augmentation target-aware can help the model explore target domain characteristics and improve the adaptation. Hence, we propose cross-domain mixture (CrDoMix) data augmentation to better condition our warm-up model to the next stage.
 
In terms of {\bf self-training} on the target domain, many works~\cite{zheng2021rectifying, zhang2019category, zhang2021prototypical, li2019bidirectional, mei2020instance, araslanov2021self} manage to optimize their self-training stage by finding proper thresholds for pseudo-labelling, which is onerous and not productive enough so far. In practice, however, self-training methods often get trapped into a performance bottleneck because the search for categorical thresholds is regarded as a trade-off between quantity and quality. Larger thresholds lead to insufficient learning, whereas smaller ones introduce noisy pseudo-labels in training. 

To handle this, we propose bilateral-consensus pseudo-supervision, a threshold-free technique which selects pseudo-labels dynamically by checking the consensus between feature-induced and probability-based labels. The feature-induced labels come from pixel-to-centroid voting, focusing on local contexts of an input. The probability-based labels from the warm-up model are better at capturing global and regional contexts by the design nature of semantic segmentation architectures. Hence, by checking the consensus of these two types of labels generated by different mechanisms, the obtained pseudo-labels provide reliable and comprehensive estimation of target domain labels. Thus, an efficient self-training stage is enabled, leading to substantial performance gainagainst prior arts.

%\indent Motivated by the aforementioned issues, we propose to totally abandon the traditional stage-wise training philosophy and provide a new training perspective for UDA segmentation. We introduce DiGA, a novel framework for domain adaptive semantic segmentation. It focuses on establishing a domain-generalized knowledge distillation scheme for stronger model warm-up \textbf{without blind feature alignment} and creating an effective \textbf{threshold-free} self-training stage (see Fig.\ref{fig:overall}).
\begin{figure}[t!]
\setlength{\belowcaptionskip}{-15pt}
\centering
\includegraphics[width=1.0\columnwidth,clip=true,trim=0 0 0 0]{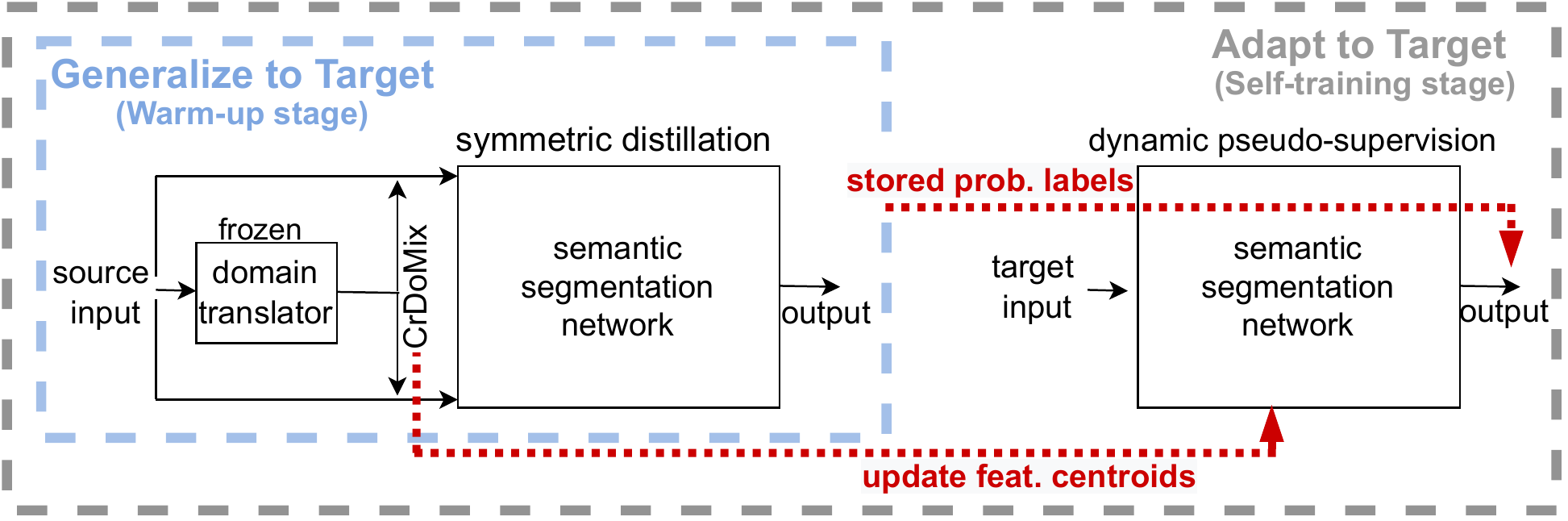}
%\caption{{\bf A systematic overview of DiGA framework}. To better support the self-training stage, combining with the proposed CrDoMix data augmentation technique, a warm-up model of higher quality is attained from our symmetric knowledge distillation approach in place of adversarial training. A threshold-free self-training stage is empowered by checking consensus between feature-induced labels and probability-based labels which are initialized from the warm-up model. CrDoMix also contributes to the update of domain-generalized class centroids to strengthen the self-training stage.}
\caption{{\bf A systematic overview of DiGA framework}. For warm-up stage, instead of aligning features between the two domains from the beginning, we propose to first make the model generalizable to an unseen domain through our symmetric distillation scheme, which can be achieved even without access to the target domain data. Coupled with our target-aware CrDoMix data augmentation technique, a warm-up model of higher quality can be obtained. To make the model better adapted to the target domain, a threshold-free self-training stage is empowered by checking the consensus between feature-induced labels and probability-based labels that are initialized from the warm-up model. CrDoMix also contributes to the update of domain-generalized class centroids to strengthen the self-training stage.}
 \label{fig:overall}
%\vspace{-6mm}
\end{figure}
In this work, we present {\bf DiGA}, a novel framework for domain adaptive semantic segmentation (see Fig.\ref{fig:overall}). Our contributions can be summarized as follows:
\begin{itemize}[noitemsep,nolistsep]
\item We introduce pixel-wise symmetric knowledge distillation sorely on source domain, which results in a stronger warm-up model and turns out to be a better option than its adversarial counterpart.
\item We introduce cross-domain mixture (CrDoMix), a novel data augmentation technique that brings further improvement to our warm-up model performance; 
\item We propose bilateral-consensus pseudo-supervision, empowering efficient self-training while abandoning categorical thresholds; 
\item Our method achieves remarkable and consistent performance gain over prior arts on popular benchmarks, \textit{e.g.}, GTA5- and Synthia-to-Cityscapes adaptation.

\end{itemize}

\section{Related Work}
\label{related_work}
\noindent {\bf Adversarial Domain Adaptation}
Adversarial learning~\cite{lowd2005adversarial,ganin2015unsupervised,tzeng2017adversarial,pei2018multi,cao2018partial} on UDA segmentation~\cite{tsai2018learning} is usually performed in GAN~\cite{goodfellow2014generative,arjovsky2017wasserstein,mao2017least,jolicoeur2018relativistic,karras2019style} fashion with a convolutional discriminator to force the output structure of the target segmentation maps to look like those from the source domain. However, adversarial training requires both source and target domain data as input and usually ends up with limited initial improvements due to blind feature alignment. In our work, in order to avoid this blind alignment, we propose pixel-wise symmetric knowledge distillation on source domain, which attains much better model generalization on target data even without observing them in training.

\noindent{\bf Knowledge Distillation}
Knowledge distillation technique~\cite{hinton2015distilling,yim2017gift,liu2019structured,wang2020intra,wang2021embracing} is first developed to perform model compression~\cite{sau2016deep,bucilua2006model} tasks. In conventional knowledge distillation pipeline, the student network learns, based on the same input, to mimic the soft output of a pretrained and fixed teacher. Therefore, the learning is unidirectional as the student has no feedback to the teacher. In our work, however, we point out that sufficient teacher-student interaction is useful in the UDA setting. We present a symmetric distillation approach to enhance the teacher-student collaboration and make the model domain generalizable.\\%From a UDA perspective, ~\cite{choi2019self,tranheden2021dacs} train the student network with labelled source domain data, and on the other hand apply hard target domain labels produced by the teacher on student predictions for pseudo-supervision, gradually improving the student network performance. Taking one step further, ~\cite{araslanov2021self} adopts larger training batch size and evolves pseudo-labelling on target domain via multi-scale fusion to force augmentation consistency between the teacher and the student. %In our work, we point out that sufficient teacher-student interaction is crucial in the UDA setting. Besides tuning the segmentation model using target domain inputs, we argue that, during warm-up training, pixel-wise symmetric knowledge distillation sorely on source domain also helps to attain impressive results on the target domain even without first introducing any target pseudo-supervision.
%taking a running mean teacher, whose gradient update is disabled, to teach a student network to approximate its behavior to improve the student's performance
\noindent{\bf Domain Generalization}
Domain generalization for semantic segmentation~\cite{choi2021robustnet,kundu2021generalize,huang2021fsdr,peng2022semantic,zhao2022style} assumes the accessibility only to one labelled source domain and aims to generalize the learned model to multiple target domains. SFDA~\cite{kundu2021generalize} prepares different classification heads for different types of augmented source data, and the model generalization can be improved by assembling outputs of the heads at test time. SHADE~\cite{zhao2022style} achieves domain generalization by utilizing a pretrained source domain teacher from the previous stage and places output consistency loss on the student network in the second stage, meanwhile, a style consistency loss is applied to the student outputs on different augmented styles. Instead of using a fixed teacher in two-stage setting, we propose a simpler but more powerful end-to-end symmetric distillation strategy, where the teacher can get adaptively improved online by the student, gradually making the model domain generalizable.\\
\noindent{\bf Data Augmentation for UDA Segmentation}
Data augmentation~\cite{simard2003best,devries2017improved,devries2017dataset,zhang2017mixup,yun2019cutmix,cubuk2020randaugment,olsson2021classmix,kundu2021generalize,tranheden2021dacs,araslanov2021self}, which increases input diversity, is a promising strategy in UDA to explore out-of-source distribution and reduces the domain discrepancy at input level. Early dedications on GAN-based image domain transfer~\cite{huang2018multimodal,liu2017unsupervised,zhu2017unpaired,hoffman2018cycada,yue2019domain,chen2019crdoco} seek to train on target-like images with the same source labels to improve the model performance on the target domain. TIR~\cite{kim2020learning} adds stylized source images carrying various texture changes to prevent the segmentation network from overfitting on one specific source texture. 
However, these approaches either introduce more training burdens (\textit{i.e.}, each augmentation is an extra batch) or depend merely on augmented images without involving the equally informative source inputs. However, in our work, we leverage random masks taken from the source label map and simultaneously embed all meaningful augmentations in a single view without introducing multiple extra training batches.\\
%DACS~\cite{tranheden2021dacs} is developed on top of~\cite{olsson2021classmix}, taking class-wise domain-mixed data for training. %Yet it does not work sorely by mixture but requires target pseudo-labels to be available. In our work, we leverage random masks taken from the source label map and simultaneously combine the source basic augmented view and the target-like translation, creating diverse inter-domain effects on one single image as a final augmentation, called CrDoMix. It is visually and experimentally verified to be a meaningful component for source domain knowledge distillation in UDA.
\noindent{\bf Self-training for UDA Segmentation}
Self-training~\cite{choi2019self,du2019ssf,pan2020unsupervised} is widely used in UDA as pseudo-labels~\cite{zou2018unsupervised,zou2019confidence} can be generated for the target domain, such that it can be treated in the same fashion as source domain.  In~\cite{zou2018unsupervised,zou2019confidence,li2019bidirectional,jiang2022prototypical}, pseudo-label acquisition is based on class-wise thresholds determined by output uncertainty. However, improper thresholds often hamper the model to learn further. Seg-Uncertainty~\cite{zheng2021rectifying} presents a pseudo-label rectifying scheme according to the prediction variance, reducing the impact of thresholds, but obtains limited boost over the threshold-based arts. ProDA~\cite{zhang2021prototypical} seeks to rectify pseudo-labels softly online according to the prototypical context estimated by target domain features towards target class centroids. Nevertheless, all pixels in rectified pseudo-labels are treated equally during training, including the false labels, thus compromising the benefit of label rectification. In our work, we exploit to enhance the label quality while not sacrificing the quantity of target pseudo-labels in a threshold-free manner. 
%CPSL~\cite{li2022class} adopts a similar strategy by rectifying soft warm-up labels using the outputs of a momentum self-labelling head, therefore, the problem of failing to filter out unreliable labels still exists. 
%We propose a dynamic pseudo-label selection procedure by pixel-wisely checking the bilateral consensus between probability-based and feature-centroid voted label maps, which filters out invalid pixels from generated labels but requires no effort on setting thresholds.
%not all uncertain predictions will lead to wrong labels, and likewise, there also exist misclassifications in highly confident predictions. Hence,
%reducing the impact from less certain labels. 
%, and re-weight their pixel-wise contributions in loss computation based on activeness measure in source feature space.
%presents a stage-wise training where pseudo-label acquisition takes thresholds based on class-wise median. 
%Specifically, each pixel in stored soft predictions gets corrected by weights measured from target feature centroids.
%------------------------------------------------------------------------
% \begin{figure}[htb!]
% %\vspace{-4.5mm}
% \centering
% \includegraphics[width=0.85\columnwidth,clip=true,trim=0 0 0 0]{figures/overall_diagram.pdf}
% \caption{ A systematic overview of our proposed DiGA framework.}
% \label{fig:overall}
% %\vspace{-2.5mm}
% \end{figure}
\section{Method}
\label{sec:proposal}
This section describes our DiGA framework for domain adaptive semantic segmentation.
For the warm up stage, we introduce the pixel-wise symmetric knowledge distillation technique in Section \ref{sec:distil} and introduce our cross-domain mixture data augmentation technique in Section \ref{sec:crdomix}. Then, for the self-training stage, we describe our threshold-free pseudo-label selection mechanism in Section \ref{sec:bi_cons}. 

Our notations are shown as the following. Let $(\mathcal{X}_s, \mathcal{Y}_s)$ denote the source dataset and $x_s \in \mathcal{X}_s$ is a source RGB image with a semantic label map $y_s \in \mathcal{Y}_s$. 
And $\mathcal{X}_t$ denotes the target domain dataset where $x_t \in \mathcal{X}_t$ stands for an unlabelled training image from the target domain. The goal is to train a segmentation network that can predict the correct per-pixel label for $\mathcal{X}_t$ with the assist of $(\mathcal{X}_s, \mathcal{Y}_s)$. 

\subsection{Pixel-wise Symmetric Knowledge Distillation}
\label{sec:distil}
%Many prior UDA works~\cite{tsai2019domain,mei2020instance,wang2020classes,huang2020contextual,wang2021uncertainty,guo2021metacorrection} train semantic segmentation network by directly feeding $x_{t}$ together with $x_{s}$ and align cross-domain feature maps or output probabilities to minimize the domain gap. In this section, we argue that a proper knowledge distillation stage using source domain data $\mathcal{X}_s$ can essentially boost the model generalization and obtain impressive results on $\mathcal{X}_t$ even without the extra effort of cross-domain training. \\
%newly identify that, apart from $\mathcal{Y}_s$'s full supervision, performing pixel-wise symmetric knowledge distillation on source domain substantially boosts the generalization of the warm-up model towards the unlabelled target domain.\\
In UDA segmentation, the purpose of a desired warm-up model is to provide strong support for the next training stage. We devise a new warm-up technique, pixel-wise symmetric knowledge distillation, to enhance the generalizability of the model towards 
the unseen target domain.

%for warm-up because the model is inadequately learned at this early training phase.
%Coupled with the proposed CrDoMix augmentation technique, our warm-up method demonstrates superiority over the widely adopted strategy based on adversarial training. \\
\begin{figure}[t!]
\centering
\setlength{\belowcaptionskip}{-10pt}
\includegraphics[width=1.0\columnwidth]{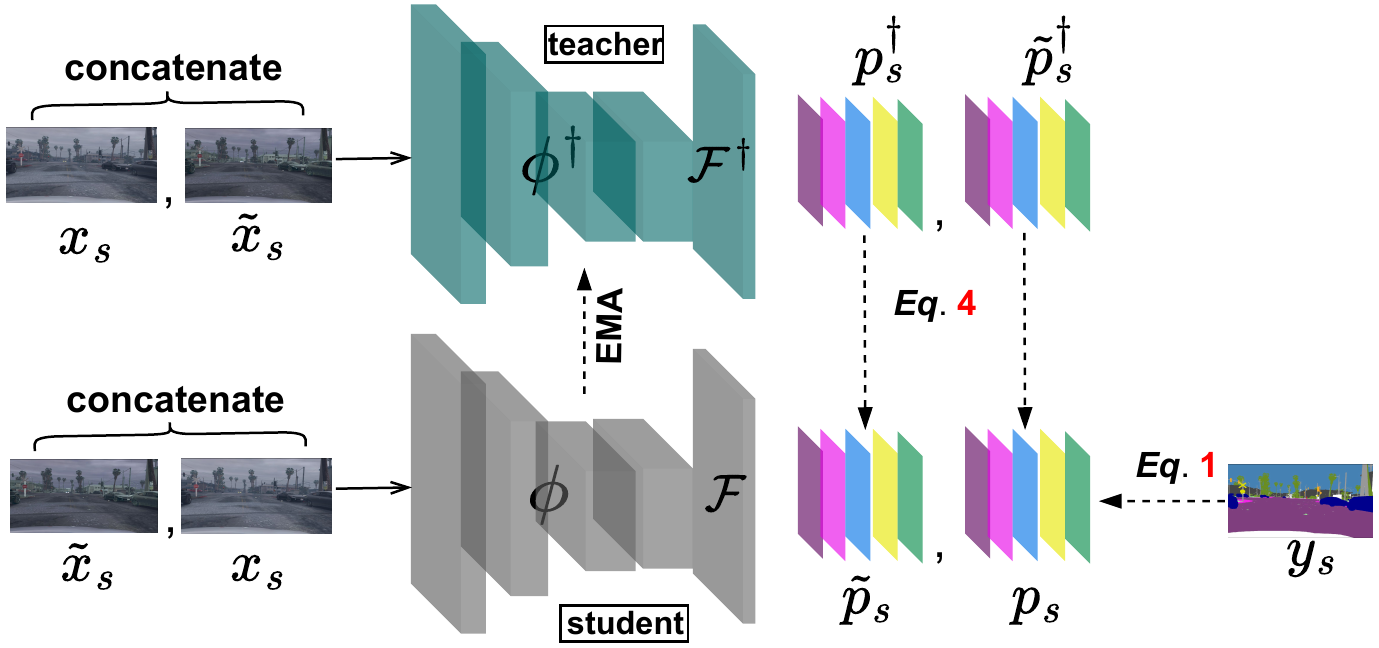}%
%\caption{The pipeline of pixel-wise symmetric knowledge distillation on the source domain.}%
%\caption{The pipeline of pixel-wise symmetric knowledge distillation. The network takes as input the fused batch of $x_{s}$ and CrDoMix X{cdm}. Distillation is only performed on the source domain.}
\caption{{\bf Overview of the warm-up stage training}. The network takes $x_{s}$ and $\Tilde{x}_{s}$ as input, then perform pixel-wise symmetric knowledge distillation sorely on source domain.}
\label{fig:model_diagram_source}
%\vspace{-5mm}
\end{figure}
\noindent {\bf Supervised Loss} As depicted in Fig.\ref{fig:model_diagram_source}, our framework consists of a teacher network $\mathcal{F}^{\dagger}\circ\phi^{\dagger}$ and a student network $\mathcal{F}\circ\phi$, where \{$\phi^{\dagger}$, $\phi$\} represent the feature encoders and \{$\mathcal{F}^{\dagger}$, $\mathcal{F}$\} stand for the semantic classifiers. Now since ground-truths for source domain are available, we first train the student network $\mathcal{F}\circ\phi$ using $\{x_{s}, y_{s}\}$ and minimize the cross-entropy loss,
\begin{align}
    \label{eq:source_seg}
    &{\mathcal{L}_{s}^{seg}} =  \sum_{h,w}\sum_{c} -{y}_{s}^{(c,h,w)} \log(p_{s})^{(c,h,w)}
\end{align}
\noindent where ${h}$, ${w}$ and $c$ are height, width and number of semantic classes respectively.\\
{\bf Distillation Loss} In parallel, we perform knowledge distillation sorely on source domain data. Based on the $x_{s}$, we first create its augmented view $\Tilde{x}_{s}$ with basic operations (such as Gaussian blur, grayscale, color jitter, etc.) Then, we introduce a pixel-wise symmetric knowledge distillation scheme which interacts different views of the input image between the teacher and the student. By exchanging cross-view information 
alternately, we want the generalization of the model to be enhanced. Specifically, we pass $x_{s}$ and $\Tilde{x}_{s}$ in a fused batch to the teacher $\mathcal{F}^{\dagger}\circ\phi^{\dagger}$ and the student $\mathcal{F}\circ\phi$ respectively to obtain the corresponding segmentation outputs (Fig.\ref{fig:model_diagram_source}),
\begin{align}
    \label{eq:prod_map}
    \{p_{s}^{\dagger},\; \Tilde{p}_{s}^{\dagger}\} = \sigma (\mathcal{F}^{\dagger}\circ\phi^{\dagger}
(\{x_{s},\; \Tilde{x}_{s}\}))\\
    \{\Tilde{p}_{s},\; p_{s}\} = \sigma (\mathcal{F}\circ\phi
(\{\Tilde{x}_{s},\; x_{s}\}))
\end{align}
\noindent where $\sigma(\cdot)$ denotes the softmax operator, whose output stands for the segmentation map in probability space. Between the teacher and student outputs, we then introduce a pixel-wise knowledge distillation loss that is computed at all pixel locations on the segmentation map. In addition, the distillation loss is symmetric, meaning that $\Tilde{p}_{s}$ is supposed to get close to $p_{s}^{\dagger}$ distribution while $p_{s}$ close to $\Tilde{p}_{s}^{\dagger}$, 
\begin{align}
    \label{eq:distil}
    {\mathcal{L}_{s}^{distil}} = \overline{\mathcal{H}(p_{s}^{\dagger},\Tilde{p}_{s})} + \alpha\overline{\mathcal{H}(\Tilde{p}_{s}^{\dagger},p_{s})}
\end{align}
where $\mathcal{H}(a,b) = -a\log(b)$ is a cross-entropy loss, and the overline indicates that the loss is computed as the mean over all pixel locations. $\alpha\in$ (0,1) is a scaling factor.\\ %Our distillation loss enhances the teacher-student collaboration, resulting in a model which generalizes well on target domain data.\\
\begin{figure}[t!]
\setlength{\belowcaptionskip}{-12pt}
\centering
\includegraphics[width=1.0\columnwidth]{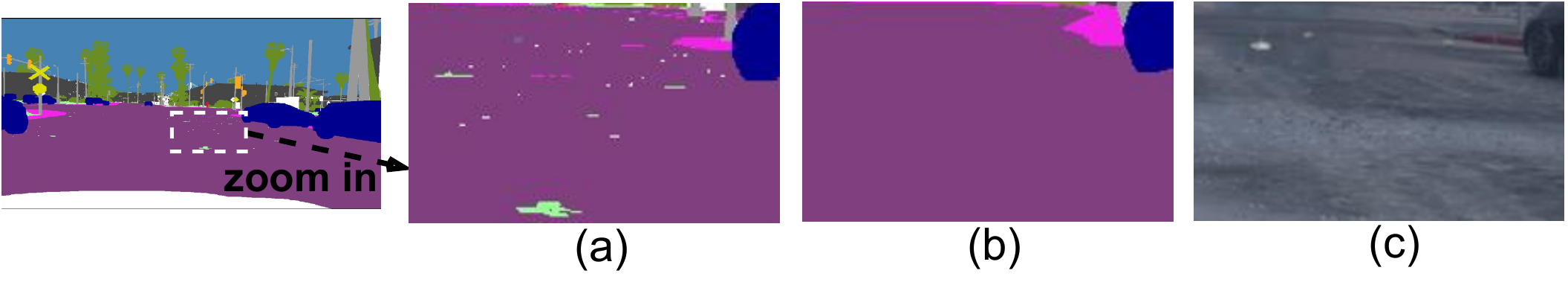}%
\caption{{\bf An illustration of the unhelpful long-tail label from the source domain}. (a) is a patch cut from source ground-truth $y_{s}$, revealing the unhelpful label pixels in it; (b) shows teacher predicted label map $\sigma (\mathcal{F}^{\dagger}\circ\phi^{\dagger}
(\Tilde{x}_{s}))$ for the same patch from $\Tilde{x}_{s}$; (c) is the corresponding RGB image for this patch. }%
\label{fig:patch}
%\vspace{-5mm}
\end{figure}
\indent Following prior arts~\cite{tranheden2021dacs,zhang2021prototypical,he2020momentum,caron2021emerging}, the gradient track of the teacher $\mathcal{F}^{\dagger}\circ\phi^{\dagger}$ is disabled, and its parameters $\Theta_{(\mathcal{F}^{\dagger}\circ\phi^{\dagger})}$ are updated per iteration according to an exponential moving average (EMA) of $\mathcal{F}\circ\phi$ with a momentum of $\xi$,
%\vspace{-0.25cm}
\begin{align}    \label{eq:param_ema}\Theta_{(\mathcal{F}^{\dagger}\circ\phi^{\dagger})} \leftarrow \xi*\Theta_{(\mathcal{F}^{\dagger}\circ\phi^{\dagger})} + (1-\xi)*\Theta_{(\mathcal{F}\circ\phi)}
\end{align}
\noindent\textbf{Why symmetric distillation works?} {\bf\textit{\Romannum{1}}}.~Distilling the soft assignments $p_{s}^{\dagger}$ to $\Tilde{p}_{s}$: By softly forcing teacher-student prediction consistency between different views (\textit{i.e.}, $x_{s}$ and $\Tilde{x}_{s}$), the network learns to produce stable outputs when input appearances are altered. This improves the model generalizability to a broader variants of source inputs and thereby, to data from an unknown domain distribution. 
{\bf\textit{\Romannum{2}}}.~Distilling the soft assignments $\Tilde{p}_{s}^{\dagger}$ to $p_{s}$: Not all source labels are helpful to learn a domain adaptable model~\cite{li2020content}. As shown in Fig.\ref{fig:patch}(a) and Fig.\ref{fig:patch}(c), the virtual source label map is much too fine-grained by definition, such that it may provide labels that are hardly recognizable in the corresponding RGB image. Given this fact, training the network with those unhelpful long-tail source label pixels will introduce bias. However, the label map derived from the teacher output (Fig.\ref{fig:patch}(b)) is smoother in those regions. Thus, introducing another symmetric distillation path on those regions of $p_{s}$ can discourage the student to learn unhelpful source labels while maintaining the correct predictions under $y_{s}$'s supervision.\\
\subsection{Cross-domain Mixture Data Augmentation} 
\label{sec:crdomix} 
\begin{figure}[t!]
\setlength{\belowcaptionskip}{-15pt}
\centering
\includegraphics[width=0.8\columnwidth]{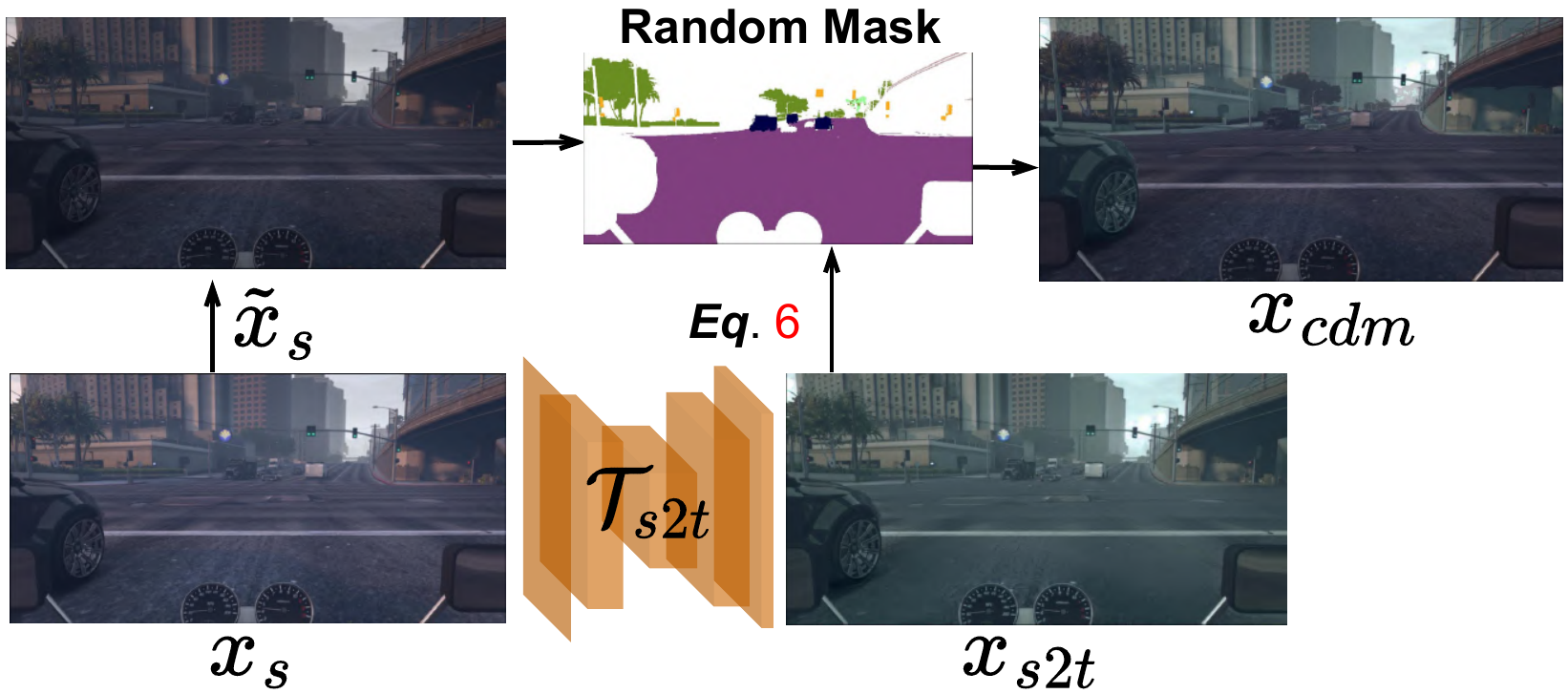}
\caption{{\bf A pictorial overview of our CrDoMix data augmentation technique}. A pretrained image domain translator is adopted.}
\label{fig:crdomix}
%\vspace{-5mm}
\end{figure}

%In the warm-up stage, the model is trained on the the labelled source domain data $\{\mathcal{X}_s, \mathcal{Y}_s\}$ and we want the model to be ready for generalizing to the target domain.
%To take full advantage of labelled source domain dataset $\{\mathcal{X}_s, \mathcal{Y}_s\}$ to train a robust semantic segmentation network, applying data augmentation on $\mathcal{X}_s$ provides an efficient solution in UDA.
%However, involving multiple augmented versions of $x_s$ at the same time in a single training iteration is impractical,since a larger batch size increases memory consumption. 
Other than basic data augmentation operations used in Sec.~\ref{sec:distil}, the domain transferred images generated by neural networks provide a supplementary way of performing data augmentation as they provide useful information across domains. Therefore, we introduce cross-domain mixture (CrDoMix) data augmentation to create a target-aware novel view out of $x_s$ (shown in Fig.\ref{fig:crdomix}). Specifically, we take a \textbf{pretrained} and \textbf{fixed} source-to-target image translator $\mathcal{T}_{s2t}$ that is trained under CycleGAN~\cite{zhu2017unpaired} pipeline, but adding a semantic edge reconstruction loss when input comes from the source domain (detailed in Supplementary), and we pass $x_s$ to $\mathcal{T}_{s2t}$ to generate its target-like version $x_{s2t} = \mathcal{T}_{s2t}({x}_{s})$. 
Then, according to the source label map $y_{s}$, we randomly select half of its available classes $c_{rs}$ and obtain a \textbf{binary mask} $\mathcal{M}$, with which we create CrDoMix augmentation $x_{cdm}$ combining $\Tilde{x}_{s}$ and $x_{s2t}$,
%\begin{align}
%    \label{eq:crdomix}
%    x_{s2t} = \mathcal{T}_{s2t}({x}_{s})\\
%    x_{cdm} = \Tilde{x}_{s}\odot \underset{c\in \mathcal{C}_{rs}}{\mathcal{M}} + %{x}_{s2t}\odot (1 - \underset{c\in \mathcal{C}_{rs}}{\mathcal{M}})
%\end{align}
\begin{align}
    \label{eq:crdomix}
        x_{cdm} = \Tilde{x}_{s} \odot \mathcal{M} + {x}_{s2t}\odot (1 - \mathcal{M})
\end{align}
\noindent where $\odot$ is element-wise multiplication. The mixture of two images using CrDoMix simultaneously embeds diverse inter-domain effects on every single augmentation $x_{cdm}$ without increasing the training batch size to cover all types of augmented images. Additionally, $x_{cdm}$ does not break the geometric structure of the original input $x_{s}$, and the domain changes that happen around instance contours can randomly appear across the whole image. Inserting target-like appearances into data augmentation, CrDoMix helps to learn cross-domain knowledge and increases the target-awareness during warm-up training. Hence, we take $x_{cdm}$ instead of just $\Tilde{x}_{s}$ for data augmentation in warm-up stage.
%\subsection{Pseudo-label selection via bilateral consensus}
\subsection{Threshold-free Self-training}
\label{sec:bi_cons}
\begin{figure}[t!]
\centering
\setlength{\belowcaptionskip}{-12pt}
\includegraphics[width=1.0\columnwidth]{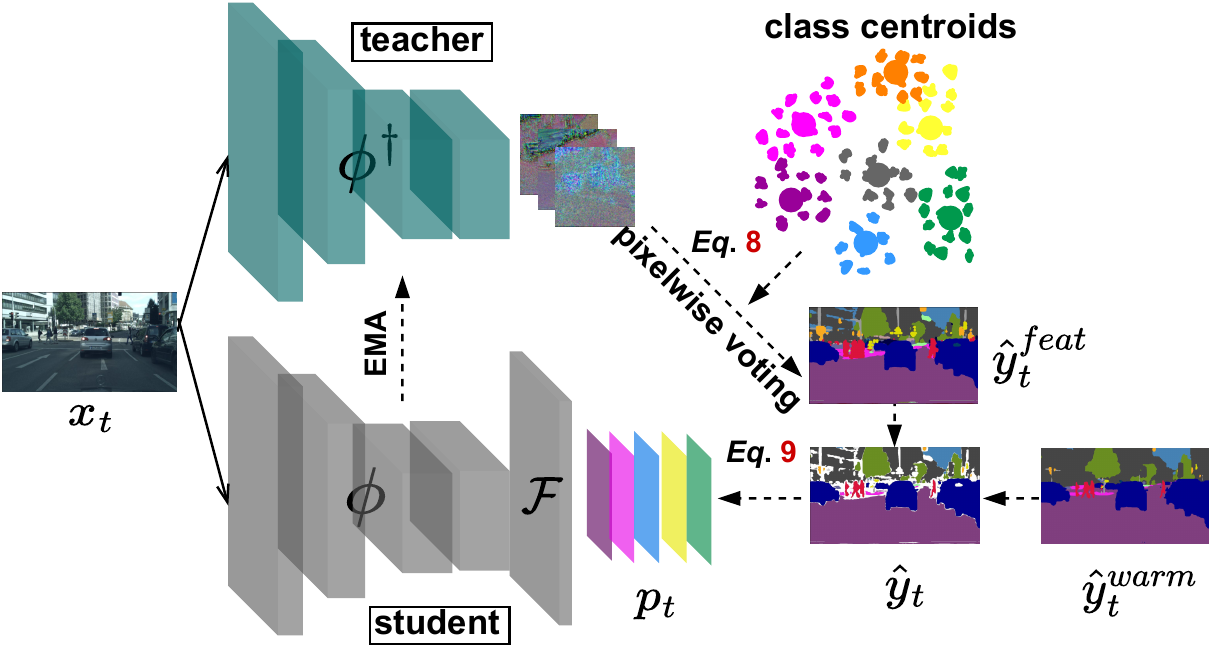}%
\caption{{\bf Overview of the proposed bilateral-consensus pseudo-supervision}. Pseudo-labels for the target domain are determined dynamically according to consensus of probability-based label maps and the ones produced by centroid-guided voting in feature space.}%
\label{fig:model_diagram_target}
%\vspace{-3.5mm}
\end{figure}

After acquiring a warm-up model that demonstrates good generalizability on target data, we then consider further adapting it to the target domain assisted by pseudo-labels in a following {\bf self-training (ST) stage}.\\   
{\bf Label Preparation} To prepare for the self-training stage, we initialize feature class centroids $\mathrm{\Lambda} = \{\rho^{(k)}, k = 1,2...,c\}$ offline using encoder $\phi$ based on our warm-up model. According to the source ground-truth we get 
$\rho^{(k)}$,
\begin{align}
    \label{eq:centroids_compute}
    &{\rho^{(k)}} =  \frac{\sum_{N_{s}}GAP(\phi(x_{cdm})^{(k)}\odot(y_{s}^{(k)}=1))}{\sum_{N_{s}}\mathbbm{1}\odot(y_{s}^{(k)}=1)}
\end{align}
where k defines the specific semantic class, $N_{s}$ is the number of images in $\mathcal{X}_{s}$, $GAP(\cdot)$ is the global average pooling operator, $x_{cdm}$ is acquired using Eq.(\ref{eq:crdomix}) to compute domain-robust centroids, and $\mathbbm{1}$ is an indicator checking whether there exists class {k} in current image $x_{cdm}$.\\
\indent In addition,  a set of label maps $\hat{y}_{t}^{warm} \in \hat{\mathcal{Y}_{t}}^{warm}$ for target domain can be obtained by processing the images from target dataset $\mathcal{X}_t$ through the warm-up model. Then, we start to consider $\mathcal{X}_t$ as input to our network for the ST stage. As shown in Fig.\ref{fig:model_diagram_target}, in each iteration, $x_{t}$ is first encoded by $\phi^{\dagger}$, and for each feature point at a specific pixel location we take feature centroids $\mathrm{\Lambda}$ as reference and vote this point to its nearest neighbor in $\mathrm{\Lambda}$. Given that $\mathrm{\Lambda}$ are initialized based on CrDoMix inputs with source labels, we can confirm that they represent the correct categorical properties of semantic classes in feature space, such that pixel-to-centroid voting can lead to reasonable target label assignments. In this way, we acquire for target domain another type of label map induced from pixel-wise voting in feature space,
%\vspace{-2mm}
\begin{align}
    \label{eq:centroids_voting}
    &{\hat{y}_{t}^{feat(jk)}} =  \mathcal{O}(\arg\min||\phi^{\dagger}(x_{t})^{(jk)}-\mathrm{\Lambda}||_{2})
\end{align}
where $\mathcal{O}$ stands for one-hot vectorization, $||\cdot||_{2}$ is L2 norm and $j \times k \in h\times w$.\\
\noindent {\bf Pseudo-Supervision} Given the challenge of searching for categorical thresholds, can efficient self-training be achieved in a threshold-free manner? Here we give this question a positive answer. On top of the source domain distillation training described in Sec.~\ref{sec:distil}, we introduce \emph{b}ilateral-consensus \emph{p}seudo-supervision (BP) with a simple but effective dynamic pseudo-label generation procedure. Considering that $\hat{y}_{t}^{feat}$ is acquired by pixel-wise voting in the feature space, it focuses more on local contexts but is likely to ignore the global semantics. For instance, $\hat{y}_{t}^{feat}$ in Fig.\ref{fig:model_diagram_target} spots the tiny objects such as traffic signs and lights, but it fails to segment building region as a whole part. On the contrary, $\hat{y}^{warm}$ obtained from probability space is able to capture global structure of an image because the design of semantic classifiers~\cite{chen2017deeplab,zhao2017pyramid,lin2017refinenet} is usually able to cover larger receptive fields rather than single pixels, such that neighboring region alignment can be better considered. Therefore, the correctness of pseudo-labels can be improved if we find pixel locations in $\hat{y}^{feat}_{t}$ and $\hat{y}^{warm}_{t}$, which are defined from different mechanisms, sharing common label pixels. To this end, we select pseudo-labels $\hat{y}_{t}$ dynamically by checking the consensus of label pixels between $\hat{y}^{feat}_{t}$ and $\hat{y}^{warm}_{t}$, and compute the self-training loss on the target domain,
\begin{align}
    %\vspace{-5mm}
    \label{eq:target_seg}
    &{\Hat{\mathcal{L}}_{t}^{seg}} =  \sum_{h,w}\sum_{c} -\hat{y}_{t}^{(c,h,w)} \log(p_{t})^{(c,h,w)}
    %\vspace{-7mm}
\end{align}
where $\hat{y}_{t}=\hat{y}^{feat}_{t}\cap\hat{y}^{warm}_{t}$. Hence, the self-training can be performed without any effort of finding class-wise thresholds. To compensate some wrong labels from warm-up models, we update $\hat{y}^{warm}_{t}$ after 50 epochs with the current student output and repeat the training till it is finished. During training, the network parameters get constantly updated, thus the class centroids are dynamically altering from the perspective of the feature encoder. Therefore, in each iteration we update the feature class centroids based on EMA,
\begin{align}
    \label{eq:centroid_ema}
    \rho^{(k)} \leftarrow \delta(\delta\rho^{(k)} + (1-\delta)\rho^{\prime\,(k)}_{s}) + (1-\delta)\rho^{\prime\,(k)}_{t}
\end{align}
where $\delta$ is the momentum and $\rho^{\prime(k)}_{s}$ is computed according to the latest $x_{cdm}$. Moreover, as $\hat{y}_{t}$ becomes growingly more reliable, we also include $\rho^{\prime(k)}_{t}$ for obtaining domain-generalized centroids.\\ 
{\bf Full Objective} So far, the full loss for training our DiGA framework can be summarized as following,
\begin{align}
    \label{eq:full_loss}
     \mathcal{L}_{DiGA} =  \lambda_{s}^{distil}\mathcal{L}_{s}^{distil} + \lambda^{seg}(\mathcal{L}_{s}^{seg}+\Hat{\mathcal{L}}_{t}^{seg})
\end{align}

\begin{figure*}[t!]
\centering
%\vspace{-0.5cm}
\includegraphics[width=2.0\columnwidth,clip=true,trim=0 0 0 0]{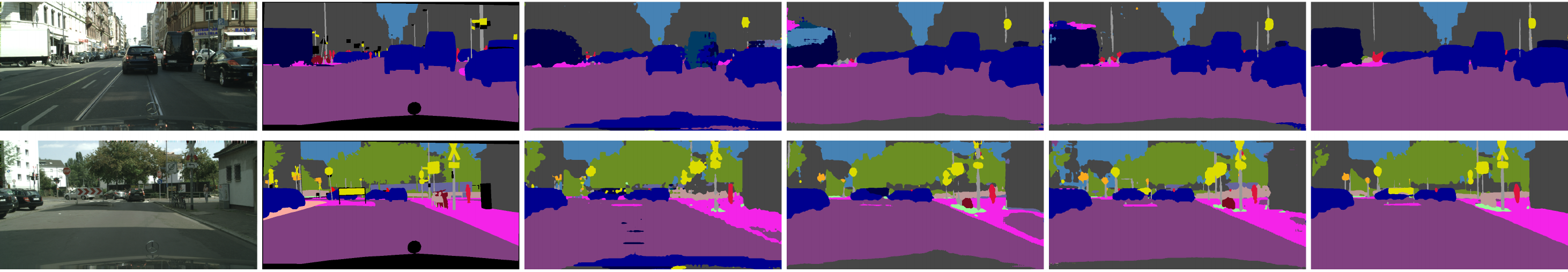}
\caption{{\bf Qualitative results of GTA5-to-Cityscapes adaptation on Cityscapes validation set}. Columns from left to right are: target domain inputs; ground-truth labels; segmentation predictions of BDL~\cite{li2019bidirectional}, ProDA~\cite{zhang2021prototypical}, CPSL~\cite{li2022class} and DiGA ({\tt ResNet}).}
\label{fig:seg_output}
%\vspace{-1.5mm}
\end{figure*}
%\vspace{-7.5mm}

\begin{table*}[t!]
    \centering
    %\vspace{-0.1cm}
    \small
    \setlength{\tabcolsep}{3pt}
    \setlength{\belowcaptionskip}{-10pt}
    \fontsize{8}{12}\selectfont
    \begin{tabular}{c|ccccccccccccccccccc|c}
    \hline Method & \rotatebox{90}{road} & \rotatebox{90}{sdwk} & \rotatebox{90}{bldng} & \rotatebox{90}{wall} & \rotatebox{90}{fence} & \rotatebox{90}{pole} & \rotatebox{90}{light} & \rotatebox{90}{sign} & \rotatebox{90}{veg} & \rotatebox{90}{trrn} & \rotatebox{90}{sky} & \rotatebox{90}{psn} & \rotatebox{90}{rider} & \rotatebox{90}{car} & \rotatebox{90}{truck} & \rotatebox{90}{bus} & \rotatebox{90}{train} & \rotatebox{90}{moto} & \rotatebox{90}{bike} & mIoU  \\ 
    \hline
    \\[-1.6em]
    %Source-only & 83.2 & 35.7 & 81.3& 29.3& 20.8& 27.4& 26.9& 19.4& 81.9& 32.2& 76.6& 51.8& 15.0& 71.9&  22.5&  28.6& 4.0& 18.5& 0.0&38.3 \\
    %\\[-1.8em]
    BDL~\cite{li2019bidirectional} &  91.0&  44.7& 84.2& 34.6& 27.6& 30.2& 36.0& 36.0& 85.0& \underline{43.6}& 83.0& 58.6& 31.6& 83.3&  35.3&  49.7& 3.3& 28.8& 35.6& 48.5\\
    \\[-1.8em]
    %CAG~\cite{zhang2019category}  &  90.4&  51.6& 83.8& 34.2& 27.8& 38.4& 25.3& 48.4& 85.4& 38.2& 78.1& 58.6& 34.6& 84.7&  21.9&  42.7& \underline{41.1}& 29.3& 37.2& 50.2\\
    %\\[-1.8em]
    %CrCDA~\cite{huang2020contextual}&  92.4&  55.3& 82.3& 31.2& 29.1& 32.5& 33.2& 35.6& 83.5& 34.8& 84.2& 58.9& 32.2& 84.7&  40.6&  46.1& 2.1& 31.1& 32.7& 48.6\\
    %\\[-1.8em]
    %FADA~\cite{wang2020classes} &  91.0&  50.6& 86.0& 43.4& 29.8& 36.8& 43.4& 25.0& 86.8& 38.3& 87.4& 64.0& 38.0& 85.2&  31.6&  46.1& 6.5& 25.4& 37.1& 50.1\\
    %\\[-1.8em]
    % Seg-Uncer~\cite{zheng2021rectifying} &  90.4&  31.2& 85.1& 36.9& 25.6& 37.5& 48.8& 48.5& 85.3& 34.8& 81.1& 64.4& 36.8& 86.3&  34.9&  52.2& 1.7& 29.0& 44.6&50.3 \\
    % \\[-1.8em]
    %DACS~\cite{tranheden2021dacs} &  89.9&  39.7& 87.9& 30.7& 39.5& 38.5& 46.4& 52.8& 88.0& \underline{44.0}& \underline{88.8}& 67.2& 35.8& 84.5&  45.7&  50.2& 0.0& 27.3& 34.0&52.1 \\
    %\\[-1.8em]
    %IAST~\cite{mei2020instance} &  94.1&  58.8 & 85.4& 39.7& 29.2& 25.1& 43.1& 34.2& 84.8& 34.6& 88.7& 62.7& 30.3& 87.6&  42.3&  50.3& 24.7& 35.2& 40.2& 52.2\\
    %\\[-1.8em]
    %UncerDA~\cite{wang2021uncertainty} &  90.5&  38.7& 86.5& 41.1& 32.9& 40.5& 48.2& 42.1& 86.5& 36.8& 84.2& 64.5& 38.1& 87.2&  34.8&  50.4& 0.2& 41.8& 54.6 & 52.6\\
    % CDGA~\cite{kim2021cross} &  91.1&  52.8& 84.6& 32.0& 27.1& 33.8& 38.4& 40.3& 84.6& 42.8& 85.0& 64.2& 36.5& 87.3&  44.4&  51.0& 0.0& 37.3& 44.9 & 51.5\\
    % \\[-1.8em]
    %SAC~\cite{araslanov2021self}&  90.4&  53.9& 86.6& 42.4& 27.3& \underline{45.1}& 48.5& 42.7& \underline{87.4}& 40.1& 86.1& 67.5& 29.7& 88.5& \underline{49.1}& 54.6& 9.8& 26.6& 45.3& 53.8\\
    %\\[-1.8em]
    ProDA$^{\ddagger}$~\cite{zhang2021prototypical}&  91.5&  52.4& 82.9& 42.0& \underline{35.7}& 40.0& 44.4& \underline{43.3}& {\bf 87.0}& {\bf 43.8}& 79.5& 66.5& 31.4& 86.7&  41.1&  52.5& 0.0& 45.4& 53.8& 53.7\\
    %ProDA$^{\ddagger}$~\cite{zhang2021prototypical} &  87.8&  56.0& 79.7& 46.3& \underline{44.8}& 45.6& 53.5& 53.5& \underline{88.6}& {\bf45.2}& 82.1& 70.7& 39.2& 88.8&  45.5&  59.4& 1.0& 48.9& \underline{56.4}& 57.5\\
    \\[-1.8em]
    %CPSL$^{\ddagger}$~\cite{li2022class} &  92.3&  59.9& 84.9& 45.7& 29.7& \underline{52.8}& \underline{61.5}& {\bf59.5}& 87.9& 41.5& 85.0& 73.0& 35.5& 90.4&  48.7&  {\bf73.9}& 26.3& {\bf53.8}& 53.9 & 60.8\\
    CPSL$^{\ddagger}$~\cite{li2022class} &  91.7&  \underline{52.9}& 83.6& \underline{43.0}& 32.3&\underline{43.7}& \underline{51.3}& 42.8& 85.4& 37.6& 81.1& \underline{69.5}& 30.0& 88.1&  \underline{44.1}&  \underline{59.9}& \underline{24.9}& \underline{47.2}& 48.4 & 55.7\\
    \\[-1.8em]
    ProCA~\cite{jiang2022prototypical} &  \underline{91.9}&  48.4& \underline{87.3}& 41.5& 31.8& 41.9& 47.9& 36.7& \underline{86.5}& 42.3& \underline{84.7}& 68.4& \underline{43.1}& \underline{88.1}&  39.6&  48.8& {\bf40.6}& 43.6& \underline{56.9} & \underline{56.3}\\
    \\[-1.8em]
    %BCL~\cite{lee2022bi} &   \underline{93.5}&  \underline{60.2}& \underline{88.1}& 31.1& 37.0& 41.9& \underline{54.7}& 37.8& {\bf89.9}& {\bf45.5}& {\bf89.9}& \underline{72.7}& 38.2& \underline{90.7}&  34.3&  53.2& 4.4& 47.2& \underline{58.5} & \underline{57.1}\\
    %\\[-1.8em]
    %DiGA(warm) &  90.9&  38.8.2& 86.1& 40.0& 29.9& 39.8& 40.9& 23.1& 85.9& 30.9& 83.4& 69.2& 36.0& 90.1&  47.5&  51.4& 6.3& 36.3& 29.6& 50.3\\
    %\\[-1.2em]
    %DiGA(ours(SST)) &  93.3&  56.2& 87.6& 40.6& 34.5& 49.9& 53.7& 41.9& 85.7& 30.9& 85.8& 72.3& 36.5& 90.9&  43.6&  57.3& 1.6& 44.3& 48.5& 55.5\\
    {\bf DiGA $(\mathrm{Ours, ResNet})$} &  {\bf95.6}&  {\bf67.4}& {\bf89.8}& {\bf51.6}& {\bf38.1}& {\bf52.0}& {\bf59.0}& {\bf51.5}& 86.4& 34.5& {\bf87.7}& {\bf75.6}& {\bf48.8}& {\bf92.5}&  {\bf66.5}&  {\bf63.8}& 19.7& {\bf49.6}& {\bf61.6}& {\bf62.7}\\
    \\[-1.6em]
    \hline
    {\bf DiGA $(\mathrm{Ours, HRNet})$} &  95.2&  65.2& 90.7& 59.0& 57.1& 57.8& 63.3& 54.8& 90.0& 42.4& 89.0& 76.8& 49.6& 91.6&  66.8&  69.8& 59.7& 24.0& 51.9& 66.1\\
    \\[-1.6em]
    \hline
    DAFormer~\cite{hoyer2022daformer} &  95.7&  70.2& 89.4& 53.5& {\bf48.1}& 49.6& 55.8& 59.4& 89.9& 47.9& {\bf92.5}& 72.2& 44.7& 92.3&  74.5&  78.2& 65.1& 55.9& 61.8& 68.3\\
    \\[-1.8em]
    {\bf DiGA $(\mathrm{Ours + DAFormer})$} &  95.7&  {\bf70.4}& {\bf89.8}& {\bf54.8}& 47.8& {\bf51.3}& {\bf57.8}& {\bf63.9}& {\bf90.3}& {\bf48.8}& 91.8& {\bf73.1}& {\bf46.6}& {\bf92.6}&  {\bf78.5}&  {\bf81.3}& {\bf74.8}& {\bf57.3}& {\bf63.2}& {\bf70.0}\\
    \\[-1.6em]
    \hline
    HRDA~\cite{hoyer2022hrda} &  96.4&  74.4& 91.0& {\bf61.6}& 51.5& {\bf57.1}& 63.9& 69.3& 91.3& 48.4& {\bf94.2}& 79.0& 52.9& {\bf93.9}&  {\bf84.1}&  85.7& 75.9& {\bf63.9}& {\bf67.5}& 73.8\\
    \\[-1.8em]
    {\bf DiGA $(\mathrm{Ours + HRDA})$} &  {\bf97.0}&  {\bf78.6}& {\bf91.3}& 60.8& {\bf56.7}& 56.5& {\bf64.4}& {\bf69.9}& {\bf91.5}& {\bf50.8}& 93.7& {\bf79.2}& {\bf55.2}& 93.7&  78.3&  {\bf86.9}& {\bf77.8}& 63.7& 65.8& {\bf74.3}\\
    \hline
    %\\[-1.8em]
    %ProDA$^{\ddagger}$~\cite{zhang2021prototypical} &  87.8&  56.0& 79.7& 46.3& {\bf44.8}& 45.6& 53.5& {\bf53.5}& {\bf88.6}& {\bf45.2}& 82.1& 70.7& \underline{39.2}& 88.8&  45.5&  59.4& 1.0& {\bf48.9}& {\bf56.4}& 57.5\\
    %\\[-1.8em]
    %DiGA$^{\ddagger}(\mathrm{ours})$ &  \underline{94.0}&  \underline{58.6}& {\bf89.0}& {\bf50.3}&  \underline{41.7}& \underline{47.7}& {\bf56.7}& 42.6& \underline{88.0}& 40.3& {\bf88.8}& {\bf72.5}& {\bf42.4}& {\bf92.2}& {\bf65.9}& {\bf62.8} &9.7 & 39.8& 46.2&{\bf59.4} \\\hline
\end{tabular}
%\vspace{1.25mm}
\caption{{\bf GTA5-to-Cityscapes adaptation results}. We compare our model performance with state-of-the-art methods. In all tables of Sec.~\ref{sec:experiments}, bold stands for {\bf best} and underline for \underline{second-best}. $\ddagger$ for fair comparison, we use their reported results after ST stage.}
\label{tab:gta5tocity}
%\vspace{-1.5mm}
\end{table*}
%means an extra distillation stage on target domain using {\tt SimCLRv2} backbone, without which the reported mIoU are 53.7 for ProDA and 55.7 for CPSL after ST stage.
\section{Experiment and Discussion}
\subsection{Datasets and Implementation Details}
For the source domain, we adopt the GTA5 dataset~\cite{richter2016playing} consisting of 24,966 images with $1914\times 1052$ resolution taken from the game engine, and SYNTHIA-RAND-CITYSCAPES dataset~\cite{ros2016synthia} composed of 9,400 images of $1280\times 760$ resolution with fine-grained segmentation labels. We adopt as target domain the 2975 urban scene training images in Cityscapes dataset~\cite{cordts2016cityscapes} and its labelled validation set containing 500 images.

We implement DiGA on an NVIDIA Quadro RTX 8000 with 48 GB memory. For fair comparison we first use ImageNet~\cite{deng2009imagenet} pretrained {\tt ResNet-101}~\cite{he2016deep} as backbone feature extractor and adopt {\tt Deeplab-V2}~\cite{chen2017deeplab} for semantic segmentation. However, to test the architectural generalizability of our method, we also train on {\tt OCRNet}~\cite{yuan2020object} with {\tt HRNet-W48}~\cite{wang2020deep} backbone, as well as on the transformer-based architectures DAFormer~\cite{hoyer2022daformer} and HRDA~\cite{hoyer2022hrda}. For the transformer-based approaches, in order to align with the end-to-end training pipeline, we implement our $\mathcal{L}_{s}^{distil}$ as an plug-and-play module, train distillation for $1$ epoch and then adapt the model to target domain with self-training.  During training, our input image size is $896\times 512$ for both domains and batch size is $3$ (containing both low and high resolution images). We augment the training data considering color jitter, color transfer~\cite{reinhard2001color}, classmix~\cite{olsson2021classmix,tranheden2021dacs}, grayscale and gaussian blur, etc.
We use the SGD~\cite{robbins1951stochastic} optimizer with a default learning rate of $2.5\times 10^{-4}$ for {\tt ResNet} but $1\times 10^{-3}$ for {\tt HRNet} setting to train our network, and for transformer-based architectures we use AdamW~\cite{loshchilov2017decoupled} optimizer with a learning rate of $6\times 10^{-5}$ following~\cite{hoyer2022daformer,hoyer2022hrda}. We use momentum $\xi$ = $\delta$ = 0.999 for EMA. During training, the losses are weighted by different hyperparameters, we set $\lambda^{seg}$ = 1, $\alpha$ = 0.5, and $\lambda_{s}^{distil}$ = 0.5 for warm-up but 0.25 for ST stage as the model relies more on self-training in this stage. We report results based on multi-scale testing (MST) as mentioned in~\cite{mei2020instance,wang2020classes,jiang2022prototypical}.%To evaluate on Cityscapes validation set we upsample the segmentation predictions to the full resolution, and we report results based on multi-scale testing (MST) as mentioned in~\cite{mei2020instance,wang2020classes,jiang2022prototypical}.

\subsection{Evaluation on Benchmark Datasets}
\label{sec:experiments}
We compare our model with state-of-the-art approaches for domain adaptive semantic segmentation. As shown in Table~\ref{tab:gta5tocity}, our DiGA framework shows leading performance among the state-of-the-art methods on GTA5-to-Cityscapes adaptation, achieving $62.7$ mIoU after the ST stage. Note that our model still outperforms ProDA~\cite{zhang2021prototypical} and CPSL~\cite{li2022class} even after they apply their proposed extra {\tt SimCLRv2}~\cite{chen2020big} distillation stage.  Our model demonstrates superior results on many important classes (e.g., road, traffic light, traffic sign, person, rider, car, truck, bus, bike etc.). A visual impression of segmentation examples generated by DiGA compared to other methods is shown in Fig.\ref{fig:seg_output}. On Synthia-to-Cityscapes adaptation in Table~\ref{tab:synthiatocity}, we also achieved state-of-the-art results (60.2 mIoU), outperforming other methods in segmenting sidewalk, traffic light as well as vehicles such as car and bicycle etc. In addition, the efficacy of our approach can be extended on more advanced architectures, reaching new milestones on both benchmarks, \textit{e.g.}, 66.1 ({\tt HRNet}) and 74.3 (HRDA) mIoU for GTA5-to-Cityscapes adaptation, as well as 62.8 ({\tt HRNet}) and 66.2 (HRDA) mIoU for Synthia-to-Cityscapes adaptation.
%Training on {\tt OCRNet}~\cite{yuan2020object} with {\tt HRNet-W48}~\cite{wang2020deep} backbone helps DiGA reach new milestones on both datasets, \textit{i.e.}, 66.1 mIoU for GTA5-to-Cityscapes adaptation and 62.8 mIoU for Synthia-to-Cityscapes adaptation.
\begin{table*}[t!]
  \centering
\small
\setlength{\belowcaptionskip}{-10pt}
\setlength{\tabcolsep}{3pt}
\fontsize{8}{12}\selectfont
\begin{tabular}{c|cccccccccccccccc|cc}
\hline Method & \rotatebox{90}{road} & \rotatebox{90}{sdwk} & \rotatebox{90}{bldng} & \rotatebox{90}{$\textrm{wall}^\star$} & \rotatebox{90}{$\textrm{fence}^\star$} & \rotatebox{90}{$\textrm{pole}^\star$} & \rotatebox{90}{light} & \rotatebox{90}{sign} & \rotatebox{90}{veg} & \rotatebox{90}{sky} & \rotatebox{90}{psn} & \rotatebox{90}{rider} & \rotatebox{90}{car} &  \rotatebox{90}{bus} &  \rotatebox{90}{mcycl} & \rotatebox{90}{bcycl} & mIoU & $\textrm{mIoU}^\star$  \\ \hline
    \\[-1.6em]
    %Source-only  &  65.1&  25.6& 77.1& 10.4& 0.0& 28.5& 0.0& 10.1& 76.0& 71.7& 52.2& 18.6& 69.4& 20.8&  15.2&  28.2& 35.6& 40.8\\
    %\\[-1.8em]
    BDL~\cite{li2019bidirectional} &  86.0&  46.7&80.3& -& -& -& 14.1& 11.6& 79.2& 81.3& 54.1& 27.9&  73.7&  42.2& 25.7& 45.3& -& 51.4\\
    \\[-1.8em]
    %CAG~\cite{zhang2019category}  &  84.7&  40.8& 81.7& 7.8& 0.0&  35.1& 13.3& 22.7& 84.5& 77.6& 64.2& 27.8&  80.9&  19.7& 22.7& 48.3& 44.5& 51.5\\
    %\\[-1.8em]
    %CrCDA~\cite{huang2020contextual}&  86.2&  44.9& 79.5& 8.3& 0.7& 27.8& 9.4&  11.8& 78.6& 86.5& 57.2& 26.1&  76.8& 39.9& 21.5& 32.1& 42.9& 50.0\\
    %FADA~\cite{wang2020classes} &  84.5&  40.1& 83.1& 4.8& 0.0& 34.3& 20.1& 27.2& 84.8& 84.0& 53.5& 22.6&  85.4&  43.7& 26.8& 27.8& 45.2& 52.5\\
    %\\[-1.8em]
    % Seg-Uncer~\cite{zheng2021rectifying} &  84.3& 37.7& 79.5& 5.3& 0.4& 24.9& 9.2& 8.4& 80.0& 84.1& 57.2& 23.0&  78.0&  38.1& 20.3& 36.5& 41.7& 48.9\\
    % \\[-1.8em]
    %DACS~\cite{tranheden2021dacs} &  80.6& 25.1& 81.9& 21.5& 2.9& 37.2& 22.7& 24.0& 83.7& {\bf90.8}& 67.6& 38.3&  82.9&  38.9& 28.5& 47.6& 48.3& 54.8\\
    %\\[-1.8em]
    %IAST~\cite{mei2020instance} &  81.9&  41.5& 83.3&17.7&{\bf4.6}&32.3& 30.9& 28.8& 83.4& 85.0& 65.5& 30.8& 86.5& 38.2& 33.1&  52.7&  49.8& 57.0 \\
    %\\[-1.8em]
    %UncerDA~\cite{wang2021uncertainty} &  87.6&  41.9& 83.1& 14.7& 1.7& 36.2& 31.3&19.9& 81.6& 80.6& 63.0& 21.8&  86.2&  40.7& 23.6& 53.1& 47.9& 54.9\\
    % CDGA~\cite{kim2021cross} &  {\bf90.7}&  \underline{49.5}& 84.5& -& -& -& 33.6&{\bf38.9}& 84.6& 84.6& 59.8& 33.3&  80.8&  \underline{51.5}& 37.6& 45.9& -& 54.1\\
    % \\[-1.8em]
    %SAC~\cite{araslanov2021self} &  \underline{89.3}&  47.2& 85.5&26.5&1.3&43.0& 45.5& 32.0& 87.1& \underline{89.3}& 63.6& 25.4& 86.9& 35.6& 30.4&  53.0&  52.6& 59.3 \\
    %\\[-1.8em]
    ProDA$^{\ddagger}$~\cite{zhang2021prototypical} &  87.1&  44.0&83.2&26.9& 0.0& 42.0& 45.8& \underline{34.2}& \underline{86.7}& 81.3& 68.4& 22.1&  87.7&  50.0& 31.4& 38.6& 51.9& 58.5\\
    %ProDA$^{\ddagger}$~\cite{zhang2021prototypical} &  87.8&  45.7&84.6&{\bf37.1}& 0.6& 44.0& \underline{54.6}& 37.0& \underline{88.1}& 84.4& 74.2& 24.3&   88.2&  51.1& 40.5& 45.6& 55.5& 62.0\\
    \\[-1.8em]
    %CPSL$^{\ddagger}$~\cite{li2022class} &  87.2& 43.9& 85.5& \underline{33.6}& 0.3& 47.7& 57.4& \underline{37.2}& 87.8& 88.5& \underline{79.0}& 32.0&  \underline{90.6}&  49.4& {\bf50.8}& 59.8& 57.9& 65.3\\
    CPSL$^{\ddagger}$~\cite{li2022class} &  87.3& 44.4& 83.8& 25.0& 0.4& \underline{42.9}& \underline{47.5}& 32.4& 86.5& 83.3& 69.6& \underline{29.1}&  \underline{89.4}&  52.1& \underline{42.6}& 54.1& 54.4& 61.7\\
    \\[-1.8em]
    ProCA~\cite{jiang2022prototypical} &  \underline{90.5}&  \underline{52.1}& 84.6&{\bf29.2}&{\bf3.3}&40.3& 37.4& 27.3& 86.4& {\bf85.9}& \underline{69.8}& 28.7& 88.7& \underline{53.7}& 14.8&  \underline{54.8}&  53.0& 59.6 \\
    \\[-1.8em]
   % BCL~\cite{lee2022bi} &  83.8&  42.2& \underline{85.3}&16.4&{\bf5.7}&\underline{43.1}& \underline{48.3}& 30.2& {\bf89.3}& \underline{92.1}& 68.2& {\bf43.1}& \underline{89.7}& 47.2& 42.2&  54.2&  \underline{55.6}& \underline{62.9} \\
    %\\[-1.8em]
    %DiGA(warm) & 87.4& 46.0& 81.0& 9.6& 0.0& 36.6& 20.4& 13.1&  84.4& 86.9& 64.8&  25.4&  88.7& 53.4&25.2 & 38.8&47.6&- \\
    %DiGA(ours(SST)) & 88.3& 52.0& 81.6& 5.9& 1.0& 49.8& 43.5& 28.8&  86.3& 82.6& 66.4&  25.7&  90.6& 62.5&30.9 & 55.5&53.2&61.1 \\
    %DiGA (ours) & \underline{89.8}& \underline{52.4}& 85.1& 8.9& 1.4& \underline{49.1}& 41.4& 28.7&  86.6& 83.0& 66.8&  27.7&  \underline{91.2}& \underline{67.5}&35.7 & \underline{57.2}&54.5&\underline{62.5} \\
    %DiGA$^{\ddagger}(\mathrm{ours}(SST)$ & 89.8&  54.8&85.3& 17.4& 1.3& 50.9& 45.8& 33.0&  87.3& 86.1& 71.3&  32.5&  92.1& 73.4&38.9 & 60.9&57.6&65.5
    {\bf DiGA $(\mathrm{Ours, ResNet})$} & 89.1&  {\bf53.4}&{\bf86.1}& \underline{28.7}& \underline{3.0}& {\bf49.6}& {\bf50.6}& {\bf34.9}&  {\bf88.2}& \underline{84.9}& {\bf71.3}&  {\bf40.9}&  {\bf91.6}& {\bf75.1}&{\bf50.3}& {\bf65.8}&{\bf60.2}&{\bf67.9}\\
    \\[-1.6em]
    \hline
    {\bf DiGA $(\mathrm{Ours, HRNet})$} & 90.6&  56.3&87.4& 38.8& 6.4& 57.7& 59.3& 50.4&  87.9& 86.4& 76.1&  47.9&  89.0& 54.2&47.2 & 69.1&62.8&69.4
    \\\hline
    \\[-1.6em]
    DAFormer~\cite{hoyer2022daformer} & 84.5&  40.7&{\bf88.4}& 41.5& 6.5& 50.0& 55.0& {\bf54.6}&  86.0& 89.8& 73.2&  48.2&  87.2& 53.2&53.9 & 61.7&60.9&67.4\\
    \\[-1.8em]
    {\bf DiGA $(\mathrm{Ours + DAFormer})$} & {\bf85.2}&  {\bf41.4}&88.2& {\bf42.6}& {\bf7.5}& {\bf52.1}& {\bf57.5}& 47.7&  {\bf87.8}& {\bf90.8}& {\bf75.0}&  {\bf50.8}&  {\bf87.8}& {\bf58.0}&{\bf58.5} & {\bf63.0}&{\bf62.1}&{\bf68.6}
    \\\hline
    \\[-1.6em]
    HRDA~\cite{hoyer2022hrda} & 85.2&  47.7&88.8& 49.5& 4.8& {\bf57.2}& {\bf65.7}& 60.9&  85.3& 92.9& {\bf79.4}&  {\bf52.8}&  89.0& {\bf64.7}&{\bf63.9} & 64.9&65.8&72.4\\
    \\[-1.8em]
    {\bf DiGA $(\mathrm{Ours + HRDA})$} & {\bf88.5}&  {\bf49.9}&{\bf90.1}& {\bf51.4}& {\bf6.6}& 55.3& 64.8& {\bf62.7}&  {\bf88.2}& {\bf93.5}& 78.6&  51.8&  {\bf89.5}& 62.2&61.0 & {\bf65.8}&{\bf66.2}&{\bf72.8}
    \\\hline
\end{tabular}
%\vspace{1.25mm}
\caption{{\bf Synthia-to-Cityscapes adaptation results}. mIoU, ${\textrm{mIoU}^\star}$ refer to {16}-class and {13}-class experimental settings, respectively. $\ddagger$ for fair comparison, we use their reported results after ST stage following~\cite{jiang2022prototypical}.}
\label{tab:synthiatocity}
%\vspace{-8.5mm}
\end{table*}

\subsection{Ablation Study}
\label{sec"ablationstudy}
 Based on {\tt Deeplab-V2} with {\tt ResNet-101} setting, we provide ablative experiments in Table~\ref{tab:ablationstudy} to analyze the effect of each component for training our DiGA framework. Each experiment is evaluated on Cityscapes validation set to compute mIoU. Compared to the source-only approach, experiment in row~(\romannum{1}) implies that distilling soft knowledge from $p_{s}^{\dagger}$ to
\begin{table}
    \centering
    \setlength{\tabcolsep}{6.0pt}
    \fontsize{8}{12}\selectfont
    \begin{tabular}{c|ccccc|cc}
        %\hline
        Method              &   a   &   b   &       c   &       d   &       e       &   mIoU  &   $\Delta$                  \\ \hline
        Source-only         &  &       &           &           &                     &   38.3  &   \textcolor{green}{+}0.0   \\
        \\[-2.0em]
        Source-only         & \cmk  &       &           &           &                 &   38.9  &   \textcolor{green}{+}0.6   \\ \hline
        (\romannum{1})      & \cmk  & \cmk  &  &      &                &   46.7  &   \textcolor{green}{+}8.4   \\
        \\[-2.0em]
        (\romannum{2})       & \cmk  & \cmk  & \cmk &      &                &    48.9  &   \textcolor{green}{+}10.6  \\ \\[-2.0em]
        (\romannum{3})      & \cmk  & \cmk  & \cmk & \cmk &                  &   51.1  &   \textcolor{green}{+}12.8  \\
        \\[-2.0em]
        (\romannum{4})      & \cmk  & \cmk  & \cmk & \cmk  & \cmk &   62.7  &   \textcolor{green}{+}24.4  \\ \hline
    \end{tabular}
    %\vspace{-0.5\baselineskip}
    \caption{%\vspace{0.2\baselineskip}Ablation study on GTA5-to-Cityscapes adaptation for %\vspace{0.2\baselineskip} DiGA components: %\vspace{0\baselineskip}
    {\bf DiGA components}: {\bf a} $ \xrightarrow[]{}$  MST,
    {\bf b} $ \xrightarrow[]{}$  $\overline{\mathcal{H}(p_{s}^{\dagger},\Tilde{p}_{s})}$, %\vspace{0.4\baselineskip} 
     {\bf c} $ \xrightarrow[]{}$  $\overline{\mathcal{H}(\Tilde{p}_{s}^{\dagger},p_{s})}$, {\bf d} $ \xrightarrow[]{}$  CrDoMix,
     and {\bf e} $ \xrightarrow[]{}$ $\Hat{\mathcal{L}}_{t}^{seg}$.}
    \label{tab:ablationstudy}
    %\vspace{-0.75cm}
\end{table}
\noindent$\Tilde{p}_{s}$ on source domain according to the first half of Eq.(\ref{eq:distil}) can lead to a considerable performance raise by 8.4\%. On top of that, we observe that adding an additional symmetric path from $\Tilde{p}_{s}^{\dagger}$ to $p_{s}$ (the second half of Eq.(\ref{eq:distil})) to our distillation brings a performance increase by 2.2\%, reaching 48.9 mIoU (row~(\romannum{2})). Combining CrDoMix, row~(\romannum{3}) gives the full model for training our warm-up stage, obtaining an increase from 48.9 to 51.1 mIoU, which already outperforms many existing self-training based methods~\cite{choi2019self,du2019ssf,pan2020unsupervised,zou2018unsupervised}. Row (\romannum{4}) indicates that, given a domain generalized warm-up model, our proposed bilateral-consensus pseudo-supervision strategy ensures high-quality self-training, boosting the mIoU from 51.1 to 62.7.
\begin{table}[t!]
    \centering
    \setlength{\belowcaptionskip}{-12pt}
    \setlength{\tabcolsep}{3pt}
    \fontsize{8}{12}\selectfont
    \begin{tabular}{c|cccc}
        \hline
        Strategy              &  Adv.~\cite{tsai2018learning}   &    Distil.   &     Adv.~\cite{tsai2018learning}+CrDoMix   &      Distil.+CrDoMix                      \\ \hline
        mIoU    & 45.2    & \underline{48.9} & 47.3      & {\bf 51.1}  
        \\[-1.8em]
       \\ \hline
    \end{tabular}
    %\vspace{-0.5\baselineskip}
    \caption{{\bf Warm-up model comparison} between adversarial training and our knowledge distillation w/ and w/o CrDoMix.}
    \label{tab:compare_warm}
    %\vspace{-0.75cm}
\end{table}

In Table~\ref{tab:compare_warm}, we compare warm-up models trained with different configurations. It indicates that our knowledge distillation technique provides a better warm-up solution than adversarial training~\cite{tsai2018learning}, witnessing a substantial improvement no matter w/ or w/o our CrDoMix data augmentation.  

\subsection{Pseudo-labelling Comparison}
We show that our proposed \emph{b}ilateral-consensus \emph{p}seudo-supervision (BP) can be a more efficient strategy for pseudo-labelling in UDA segmentation. It assures a sufficient amount of reliable pixels in the resulting pseudo-labels but requires no effort to look for a proper threshold for each class. To verify this, we train ST stage models using different pseudo-labelling strategies, ablating on $\hat{y}^{feat}_{t}$, $\hat{y}^{warm}_{t}$ and comparing with two influential strategies in this field, BDL~\cite{li2019bidirectional} and ProDA~\cite{zhang2021prototypical}. For fair comparison, we start with our warm-up model in all experiments but only replace our BP with other strategies and evaluate the ST stage model performances on Cityscapes validation set. As shown in Table~\ref{tab:compare_mIoU}, training sorely with either $\hat{y}^{feat}_{t}$ or $\hat{y}^{warm}_{t}$ fails to show advantageous results (Exp.({1})\;\&\;Exp.({2})), which indicates that it is required to check the consensus between $\hat{y}^{feat}_{t}$ and $\hat{y}^{warm}_{t}$ in order to achieve the best performance among all (Exp.({5})). We can also observe through Exp.({3}) and Exp.({4}) that replacing our BP strategy with BDL~\cite{li2019bidirectional} and ProDA~\cite{zhang2021prototypical} strategies will lead to lower scores, obtaining 56.2 and 59.5 mIoU, respectively. Nevertheless, the scores are still higher than those reported in their original papers (48.5 and 53.7 mIoU), which also implies the superiority of our proposed warm-up strategy, confirming that a more advanced warm-up model contributes largely to the performance gain in ST stage.\\
\begin{table}[t!]
    \centering
    %%\vspace{-2.75\baselineskip}
    %\setlength{\intextsep}{-3cm}
    \setlength{\tabcolsep}{3.5pt}
    \setlength{\belowcaptionskip}{-5pt}
    \fontsize{8}{12}\selectfont
    \begin{tabular}{c|ccccc}
        %\hline
        Strategy              &  $(1) \hat y_t^{feat}$   &   $(2) \hat y_t^{warm}$   &     $(3)$BDL   &       $(4)$ProDA   &   $(5) \hat y_t (ours)$                      \\ \hline
        mIoU    & 52.1    & 53.8 & 56.2      & 59.5           & {\bf62.7}       
        \\[-1.8em]
       \\ \hline
    \end{tabular}
    %\vspace{-0.5\baselineskip}
    \caption{mIoU comparison of applying different pseudo-labelling techniques to train ST stage \textbf{based on our warm-up model}.}
    \label{tab:compare_mIoU}
    %\vspace{-0.75cm}
\end{table}
\begin{figure}[t!]
    \centering
    \setlength{\belowcaptionskip}{-15pt}
    \includegraphics[width=\linewidth]{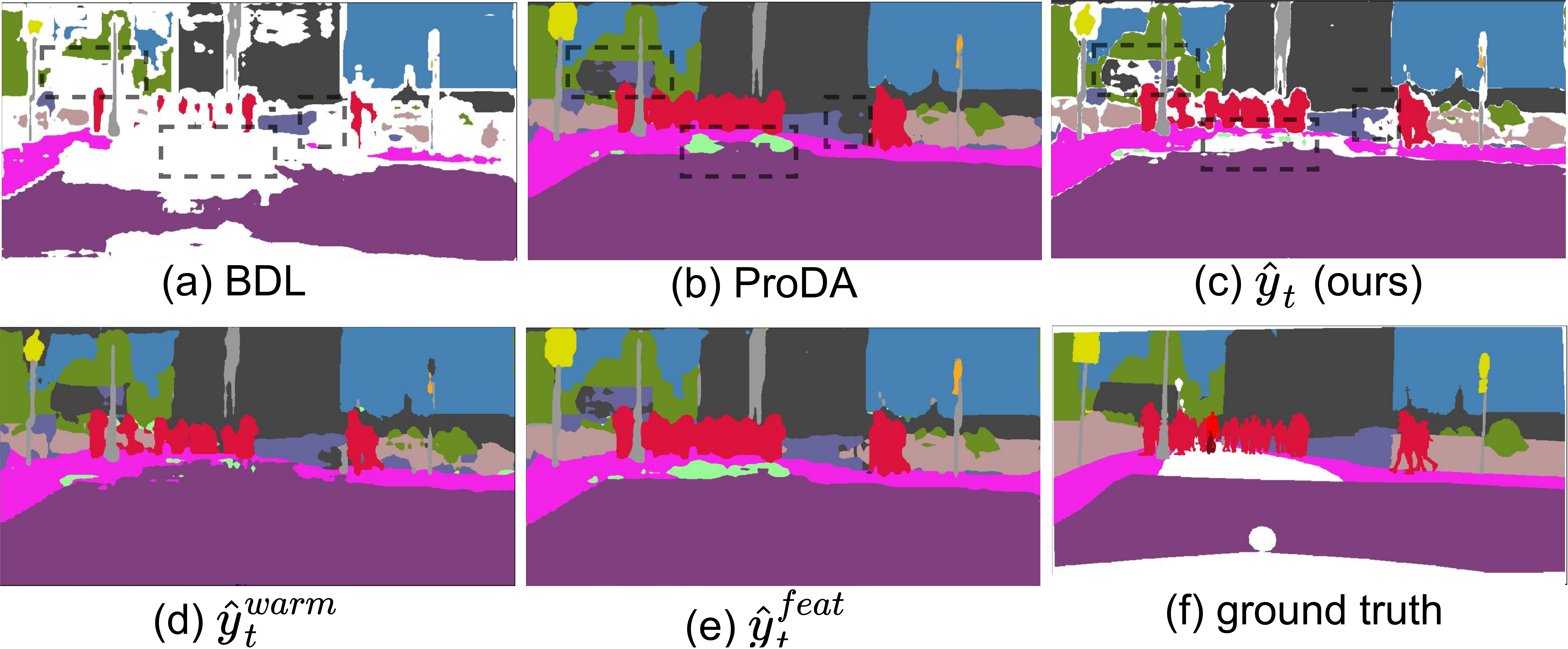}
    \caption{{\bf Comparison of different pseudo-labelling techniques} given the same input image, and ground truth (f) is only adopted for comparison. Dashed black boxes reveal the major differences. }
    \label{fig:compare_pseudo}
%\vspace{-0.35cm}
\end{figure}
A visual intuition can be obtained by comparing Fig.\ref{fig:compare_pseudo} (a), (b) and (c) with the ground-truth (f). Given the same target domain input, BDL~\cite{li2019bidirectional} manages to exclude predicted labels that are less confident than the class-wise thresholds, which misses a number of label pixels with correct predictions, leading to insufficient self-training. ProDA~\cite{zhang2021prototypical}, on the other hand, keeps a full rectified label map by element-wisely re-weighting initial soft assignments using prototype-based soft assignments. Nevertheless, ProDA has no label filtering mechanism, and the worst that can happen is that the model will be trained with wrong labels if false prototypical predictions override the correct initial ones. However, our BP strategy provides a better trade-off, showing superiority, both quantitatively and qualitatively, over the above methods in generating pseudo-labels.

\subsection{Pseudo-labelling and Prediction Uncertainty}
\label{sec:uncertainty}
\begin{figure}[t!]
    \centering 
    \setlength{\belowcaptionskip}{-3pt}
    \includegraphics[width=0.9\columnwidth]{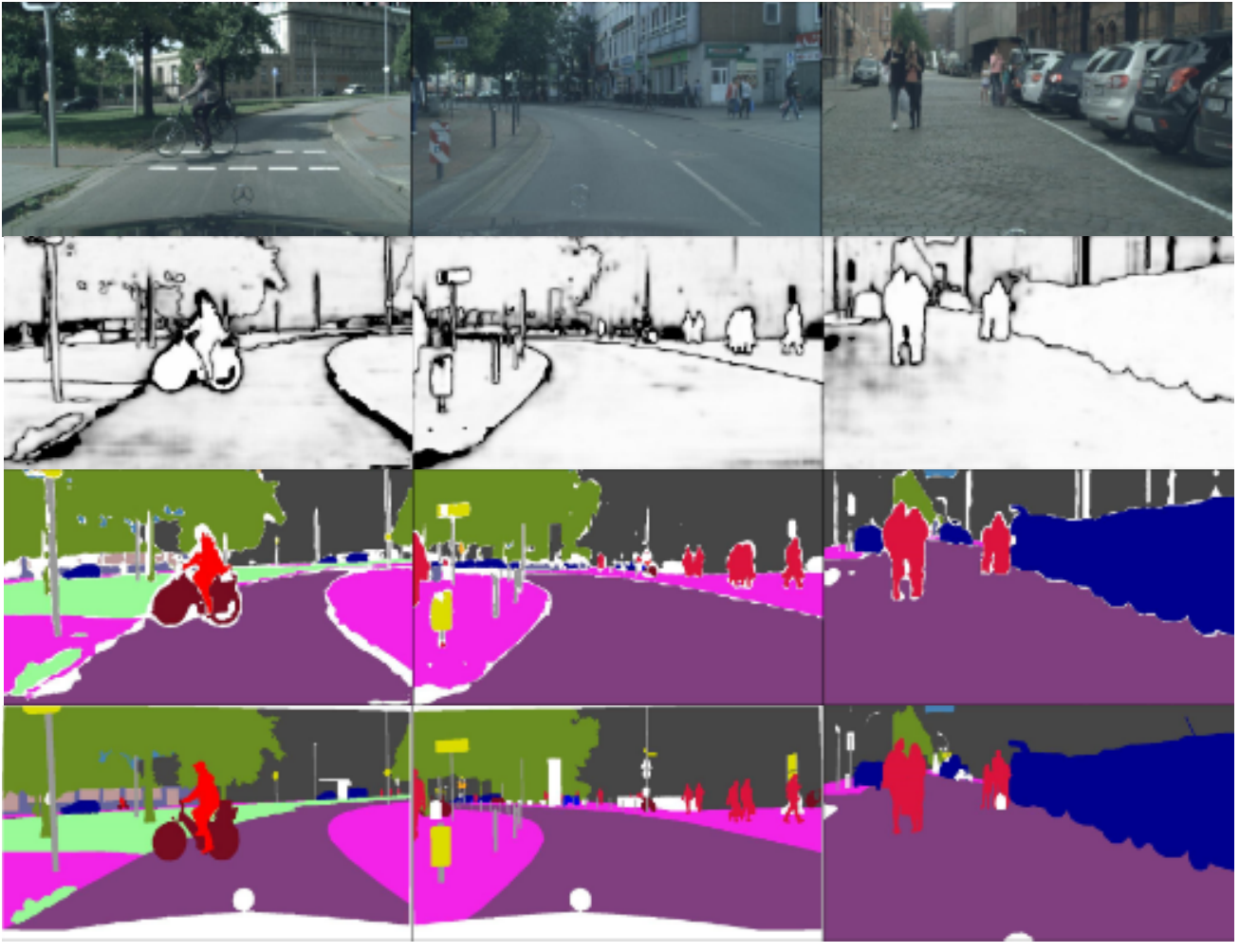}%
        \caption{{\bf Label uncertainty in ST stage}. From top to bottom: target input, prediction uncertainty (darker means lower confidence), pseudo-label, ground-truth (unused in training). }
        \label{fig:uncertainty}
\end{figure}
We reveal experimentally that our BP strategy is strongly correlated with pixel uncertainty in pseudo-label selection. During ST stage training, we visualize the model uncertainty on target inputs based on the output probabilities (See Fig.~\ref{fig:uncertainty}). For each mini-batch of data, the prediction uncertainty varies class-wisely and instance-wisely, which can hardly be described by some specific thresholds. However, we observe that BP strategy dynamically and efficiently identifies most pixel locations with relatively lower confidence ratio and successfully filters them out from the pseudo labels. Therefore, we confirm that, our BP strategy, despite of being threshold-free, can still obtain reliable pseudo-labels to improve self-training.

\subsection{Extensive Experiments}
\label{sec:extensive}
\noindent {\bf Domain Generalization} Even though we focus on solving the problem of UDA segmentation in this work, we also want to share that our warm-up strategy can offer a `free lunch' for domain generalization on semantic segmentation and achieve superior results compared to SOTA methods. Decoupling CrDoMix (no $x_{s2t}$) from our data augmentation, meaning no information of target domain is touched, our warm-up distillation method works surprisingly well on domain generalization for semantic segmentation, \textit{i.e.}, train on one labelled source domain and generalize to multiple target domains. Our model trained on GTA5~\cite{richter2016playing} dataset can be well generalized on the validation set of Cityscapes~\cite{cordts2016cityscapes}, BDD100k~\cite{yu2020bdd100k}, Mapillary~\cite{neuhold2017mapillary} and Synthia~\cite{ros2016synthia}, outperforming existing SOTA methods of domain generalizable semantic segmentation~\cite{choi2021robustnet, peng2022semantic} by considerable margins (see Table~\ref{tab:table_dg}). This \textbf{cannot} be accomplished by adversarial learning as it requires target data for training.

\noindent {\bf Semi-supervised Semantic Segmentation} As shown in Table~\ref{tab:table_semi}, following the same partition protocols on Cityscapes dataset, our stage-wise training pipeline also shows impressive performance on the task of semi-supervised semantic segmentation. In particular, the less labels available, the more advantageous DiGA is.

\begin{table}[t!]
    \centering
        \setlength{\tabcolsep}{9pt}
        \setlength{\belowcaptionskip}{-3pt}
    \fontsize{8}{12}\selectfont
    \begin{tabular}{ccccc} \Xhline{2\arrayrulewidth}
        \multirow{2}{*}{Method}    &    \multicolumn{4}{c}{Train on GTA5 (G)} \\ \cline{2-5}
            &  $\rightarrow$C  &  $\rightarrow$B    &  $\rightarrow$M    &    $\rightarrow$S      \\ \Xhline{2\arrayrulewidth}
            %Base~\cite{wang2021embracing} &   41.91      &   35.23      &   35.78      &   28.62 \\
           ISW~\cite{choi2021robustnet} &   42.87      &   38.53      &   39.05      &   29.58 \\
        SFDA~\cite{wang2021embracing} &   43.50      &   -      &   -      &   - \\
        SAN-SAW~\cite{peng2022semantic}    &   45.33      &   41.18      &   40.77      &   31.84 \\
        SHADE~\cite{zhao2022style}&   46.66      &   43.66      &   45.50      &   - \\
        \hline
        Our Distillation    &   {\bf48.87}      &   {\bf44.42}      &   {\bf51.78}      &   {\bf37.17} \\\hline
        \Xhline{2\arrayrulewidth}
    \end{tabular}
    \caption{{\bf mIoU comparison with SOTA methods for domain generalization}. G, C, B, M and S denote GTA5, Cityscapes, BDD100k, Mapillary and Synthia, respectively. For fair comparison, all the listed methods are based on {\tt ResNet-101} backbone.} %Base is reproduced from~\cite{wang2021embracing}, a two-stage distillation method.}
    \label{tab:table_dg}
\end{table}

\begin{table}[t!]
    \centering
    \setlength{\belowcaptionskip}{-10pt}
    \setlength{\tabcolsep}{5pt}
    \fontsize{8}{12}\selectfont
    \begin{tabular}{c|c|c|c|c} \hline %\Xhline{2\arrayrulewidth}
        \multirow{2}{*}{Method}    &   \multicolumn{4}{c}{Cityscapes} \\ \cline{2-5}
                &   1/16 (186)     &   1/8 (372)     &   1/4 (744)     &   1/2 (1488)     \\ \hline\hline
        CPS~\cite{chen2021semi}    &   75.09      &   77.92      &   79.24      &   80.67 \\
        Ours    &   {\bf76.86}      &   {\bf78.51}      &   {\bf80.01}      &   {\bf80.93} \\\hline
    \end{tabular}
    \caption{{\bf mIoU comparison of semi-supervised semantic segmentation} using {\tt HRNet} backbone, based on which SOTA performance of CPS~\cite{chen2021semi} is reported. Evaluation performed on Cityscapes validation set under different partition protocols.}
    \label{tab:table_semi}
\end{table}

\section{Conclusion}
In this work, we propose DiGA framework for domain adaptive semantic segmentation. It first enhances the model generalization to the target domain by pixel-wise symmetric knowledge distillation performed on the source dataset. Supported by this strong warm-up model, our bilateral-consensus pseudo-supervision strategy reinforces the model adaptability during self-training without thresholds, demonstrating SOTA performances on popular benchmarks. Besides, through extensive experiments we also observe the efficacy of DiGA on domain generalized and semi-supervised semantic segmentation tasks. We believe DiGA can provide a universal solution for element-wise 2D (or 3D) classification problems under UDA or semi-supervised setting, into which we will investigate more in the future.

%%%%%%%%% REFERENCES
{\small
\bibliographystyle{ieee_fullname}
\bibliography{diga}
}

\end{document}